\documentclass[conference]{ieeeconf}
\IEEEoverridecommandlockouts

\usepackage{cite}
\usepackage{relsize}
\usepackage{textcomp}
\usepackage{xcolor}
\usepackage{pgfplots}
\usepackage{pgfplotstable}
\def\BibTeX{{\rm B\kern-.05em{\sc i\kern-.025em b}\kern-.08em
    T\kern-.1667em\lower.7ex\hbox{E}\kern-.125emX}}

\usepackage{pgfplots}
\usepackage{siunitx}
\let\labelindent\relax
\usepackage{enumitem}

\usepackage[T1]{fontenc} 
\usepackage[utf8]{inputenc} 
\usepackage[english]{babel} 

\usepackage{amsmath} 
\usepackage{amsfonts}
\usepackage{amssymb}  

\usepackage{booktabs} 

\usepackage{dsfont}
\usepackage{graphicx}
\usepackage{caption}
\usepackage{tikz}
\usepackage{pgfplots}
\usepackage{scalefnt}
\pgfplotsset{compat=newest}
\usepgfplotslibrary{fillbetween}
\usepgfplotslibrary{statistics}
\usepgfplotslibrary{groupplots}

\usepackage{epsfig} 
\usepackage{times} 
\usepackage{siunitx}
\usepackage{calc}
\usepackage[hyphens]{url}
\usepackage{svg}
\renewcommand{\baselinestretch}{0.859}
\usepackage{adjustbox}

\usepackage{tabularx}
\usepackage{bbm}
\usepackage{dsfont}

\usepackage{diagbox}
\usepackage{makecell}
\usepackage{multirow}

\usepackage[linesnumbered,ruled,vlined]{algorithm2e}
\usepackage[noend]{algpseudocode}

\usepackage{svg}

\usepackage{scalerel}
\usepackage{tikz}
\usetikzlibrary{svg.path}

\definecolor{orcidlogocol}{HTML}{A6CE39}
\tikzset{
	orcidlogo/.pic={
		\fill[orcidlogocol] svg{M256,128c0,70.7-57.3,128-128,128C57.3,256,0,198.7,0,128C0,57.3,57.3,0,128,0C198.7,0,256,57.3,256,128z};
		\fill[white] svg{M86.3,186.2H70.9V79.1h15.4v48.4V186.2z}
		svg{M108.9,79.1h41.6c39.6,0,57,28.3,57,53.6c0,27.5-21.5,53.6-56.8,53.6h-41.8V79.1z M124.3,172.4h24.5c34.9,0,42.9-26.5,42.9-39.7c0-21.5-13.7-39.7-43.7-39.7h-23.7V172.4z}
		svg{M88.7,56.8c0,5.5-4.5,10.1-10.1,10.1c-5.6,0-10.1-4.6-10.1-10.1c0-5.6,4.5-10.1,10.1-10.1C84.2,46.7,88.7,51.3,88.7,56.8z};
	}
}

\newcommand\orcidicon[1]{\href{https://orcid.org/#1}{\mbox{\scalerel*{
				\begin{tikzpicture}[yscale=-1,transform shape]
					\pic{orcidlogo};
				\end{tikzpicture}
			}{|}}}}

\usepackage{hyperref} 
\hypersetup{
    colorlinks=true,
    breaklinks=true,
    linkcolor=blue,
    filecolor=magenta,      
    hypertexnames=true,
    urlcolor=cyan,
    pdfpagemode=FullScreen,
    citecolor=green
    }
\title{\LARGE \textbf{Enhancing System Self-Awareness and Trust of AI:\\A Case Study in Trajectory Prediction and Planning}$^{*}$}

\author{Lars Ullrich $^{\orcidicon{0009-0001-8166-3118}}$, Zurab Mujirishvili $^{\orcidicon{0009-0006-2669-2799}}$, and Knut Graichen $^{\orcidicon{0000-0003-2865-8093}}$
	\thanks{*This research is accomplished within the project ”AUTOtech.agil” (FKZ 01IS22088Y). We acknowledge the financial support for the project by the Federal Ministry of Education and Research of Germany (BMBF).}
	\thanks{The authors are with the Chair of Automatic Control, Friedrich-Alexander-Universität Erlangen-Nürnberg (FAU), Germany {\tt\footnotesize \{lars.ullrich, zurab.mujirishvili, knut.graichen\}@fau.de}}
}

\usepackage{hologo}
\usepackage{listings}
\begin{document}

\twocolumn[
\begin{@twocolumnfalse}
\Huge {IEEE copyright notice} \\ \\
\large {\copyright\ 2025 IEEE. Personal use of this material is permitted. Permission from IEEE must be obtained for all other uses, in any current or future media, including reprinting/republishing this material for advertising or promotional purposes, creating new collective works, for resale or redistribution to servers or lists, or reuse of any copyrighted component of this work in other works.} \\ \\
	
{\Large Accepted to be published at the \emph{2025 36th IEEE Intelligent Vehicles Symposium (IV)}, Cluj-Napoca, Romania, June 22 - 25, 2025.} \\ \\
	
Cite as:
	
\vspace{0.1cm}
\noindent\fbox{%
	\parbox{\textwidth}{
		{L.~Ullrich, Z.~Mujirishvili, and K.~Graichen, "Enhancing System Self-Awareness and Trust of AI: A Case Study in Trajectory Prediction and Planning," in \emph{2025 36th IEEE Intelligent Vehicles Symposium (IV)}, to be published.}
	}%
}
\vspace{2cm}

\end{@twocolumnfalse}
]

\noindent\begin{minipage}{\textwidth}

\hologo{BibTeX}:
\footnotesize
\begin{lstlisting}[frame=single]
@inproceedings{ullrich2025trustai,
	author={Ullrich, Lars and Mujirishvili, Zurab and Graichen, Knut},
	booktitle={2025 36th IEEE Intelligent Vehicles Symposium (IV)},
	title={Enhancing System Self-Awareness and Trust of AI: A Case Study in Trajectory Prediction and Planning},
	address={Cluj-Napoca, Romania},
	year={2025},
	publisher={IEEE. to be published}
}
\end{lstlisting}
\end{minipage}

\thispagestyle{empty}
\pagestyle{empty}
\bstctlcite{IEEEexample:BSTcontrol}

\maketitle

\begin{abstract}
In the trajectory planning of automated driving, data-driven statistical artificial intelligence (AI) methods are increasingly established for predicting the emergent behavior of other road users. While these methods achieve exceptional performance in defined datasets, they usually rely on the independent and identically distributed (i.i.d.) assumption and thus tend to be vulnerable to distribution shifts that occur in the real world. In addition, these methods lack explainability due to their black box nature, which poses further challenges in terms of the approval process and social trustworthiness. Therefore, in order to use the capabilities of data-driven statistical AI methods in a reliable and trustworthy manner, the concept of TrustMHE is introduced and investigated in this paper. TrustMHE represents a complementary approach, independent of the underlying AI systems, that combines AI-driven out-of-distribution detection with control-driven moving horizon estimation (MHE) to enable not only detection and monitoring, but also intervention. The effectiveness of the proposed TrustMHE is evaluated and proven in three simulation scenarios.  
\end{abstract}
\section{Introduction}\label{introduction}
Within automated driving (AD) research, data-driven statistical artificial intelligence (AI) methods are currently showing impressive results \cite{li2022bevformer, carion2020end, shi2022motion, seff2023motionlm}. However, as they typically assume independent and identically distributed (i.i.d.) data, they often struggle with changes between development and operational data, short distribution shifts \cite{zhang2013domain, zhou2022domain}. But due to the temporal dynamics and open long-tail distribution of real world scenarios, distribution shifts are to be expected during operation. 

While modern iterative development processes are able to account for newly discovered shifts and trigger conditions, e.g., corner cases, throughout the lifecycle \cite{favaro2023building, ullrich2024expanding}, safe and reliable operation necessitates an profound handling at the time of occurrence. Even though AI methods are evolving to be more robust with respect to out-of-distribution (OOD) events, for example through OOD generalization \cite{zhou2022domain} or meta-learning strategies \cite{garnelo2018conditional, ullrich2023cnp}, this progress offers no guarantees in the case of unknown unknowns \cite{scholkopf2021toward, ullrich2024aisafety_assurance}. Yet, hybrid or purely AI-based methods often outperform classical approaches significantly, thus providing generally improved safety \cite{grigorescu2020survey, chen2023end}. As a result and to leverage the benefits, the inherently data-dependent challenges must be addressed systematically. Furthermore, as AI methods are advancing rapidly, complementary approaches that are independent of specific underlying AI techniques are currently pursued to ensure continued validity in the face of future innovations. 

A first step in this direction are OOD-detection approaches that facilitate to check the assumptions of input distributions of AI systems \cite{hendrycks2016baseline, liu2020energy}. A counterpart to this is the verification of the output, e.g. the physical feasibility of a given trajectory prediction. In addition, safety monitors have emerged in recent years that observe AI inputs and outputs to monitor the behavior of the AI system \cite{ferreira2021benchmarking, guerin2022unifying, liang2017enhancing}. It is worth noting that efforts such as monitoring are not only advisable for high-risk AI systems, but are also required by law \cite{eu_parliament_2024corr}. Despite this, the targeted use of the gained knowledge to prevent harm during operation has not yet been systematized. 

In general, high-risk AI applications--such as the prediction of other road users--are typically embedded in dynamic, real world cyber-physical systems. Consequently, transitioning from single-time-step analysis to horizon-based approaches is essential to capture temporal dependencies and ensure reliable decision-making under uncertainty. Furthermore, interpreting AI trustworthiness as AI confidence and thus as uncertainty about the reliability of the AI system provides a way of thinking that can be natively integrated into existing probabilistic approaches, which are becoming increasingly established in high-risk applications. Building on this, we propose TrustMHE, a bridging methodology between AI engineering and control engineering that estimates the reliability uncertainty of an AI system. Leveraging this uncertainty estimation enables systematic monitoring, detection, and intervention while maintaining design flexibility. 

In this paper, we aim to address key challenges of data-driven statistical AI systems to enhance system self-awareness and trust in AI, while focusing on real world applications. Accordingly, the methodology is explored and evaluated through a case study in the context of trajectory prediction and planning. The main contribution of this paper is threefold: 

\begin{enumerate}
    \item We present the TrustMHE approach, which is designed for dynamic real world applications and enables native function monitoring, degradation detection, and mitigation intervention. 
    \item We apply and evaluate the approach to the trajectory prediction of other road users as part of trajectory planning.
    \item We provide a systematic approach to overcome existing challenges of data-driven statistical AI methods to safely and reliably leverage their capabilities in the real world.
\end{enumerate}

The paper is structured as follows: the use case of trajectory prediction and planning is outlined in Section \ref{fundamentals}, representing the evaluation baseline. The methodology of TrustMHE is elaborated in Section \ref{methodology}. The evaluation is described in Section \ref{analysis} along with the corresponding results. Finally, findings are discussed and summarized in Section \ref{conclusion}.
\section{Trajectory Prediction \& Planning}\label{fundamentals}

The use case considered in this paper involves a state-of-the-art data-driven statistical AI method for trajectory prediction, which is integrated into the broader task of trajectory planning. The default use case setting without integrated TrustMHE represents the baseline of the case study and is therefore discussed in more detail below.

\subsection{Motion Transformer-Based Trajectory Prediction}
Nowadays, trajectory prediction models that forecast the evolution of nearby road agents across time horizons of up to eight seconds are predominately data-driven statistical AIs. While occupancy-based predictions have become increasingly popular recently \cite{mahjourian2022occupancy, tong2023scene}, multimodal interaction-aware trajectory predictions remain prevalent \cite{trentin2023multi, shi2024mtrpp, gan2023mgtr}. Within this setting, a set $\Gamma_{\mathrm{pre}} = \{\Gamma_{k_{\mathrm{pre}}}\}_{k_{\mathrm{pre}}=1}^{K_{\mathrm{pre}}}$ is defined, where each modality $\Gamma_{k_{\mathrm{pre}}} = \{\hat{\boldsymbol{\tau}}_{k_{\mathrm{pre}}}, c_{k_{\mathrm{pre}}}\}$ consists of a trajectory $\hat{\boldsymbol{\tau}}_{k_{\mathrm{pre}}} = \{\hat{\boldsymbol{\tau}}_{k_{\mathrm{pre}},t_{\mathrm{pre}}}\}_{t_{\mathrm{pre}}=1}^{T_{\mathrm{pre}}}$ over a time horizon of $T_{\mathrm{pre}}$, along with a scalar confidence value $c_{k_{\mathrm{pre}}}$ associated with the trajectory. The evaluation of such multimodal predictions is based on various performance metrics, such as mean Average Precision (mAP), minimum Average Displacement Error (minADE), minimum Final Displacement Error (minFDE) and the missRate (MR) \cite{ettinger2021large}.

The Motion Transformer (MTR) \cite{shi2022motion} serves as the baseline architecture in the field of multimodal trajectory prediction, achieving the first place in the Waymo Motion Prediction Challenge 2022\footnote{\url{https://waymo.com/open/challenges/2023/motion-prediction/}}. As shown in Table \ref{tab:performance_metrics}, subsequent models building on MTR have consistently taken the first place in the following years. In addition, various other models, such as \cite{liu2023transformer, kang2023iair, lin2023eda, gan2023mgtr}, have also extended the original MTR architecture, consolidating its position as a foundational framework in the field and establishing it as the basis for this work.

\begin{table}[h!]
  \centering
  \caption{Evolution of MTR-based trajectory prediction models.}
   \resizebox{\columnwidth}{!}{
  \label{tab:performance_metrics}
  \begin{tabular}{p{1.5cm}p{0.96cm}p{0.9cm}p{0.9cm}p{0.9cm}p{0.9cm}}
    \toprule
    Method & Note & mAP$\uparrow$ & minADE$\downarrow$ & minFDE$\downarrow$ & MR$\downarrow$ \\
    \midrule
    MTR \cite{shi2022motion}& 1st 2022 & 0.4129 & 0.6050 & 1.2207& 0.1351 \\
    MTR++\cite{shi2024mtrpp}& 1st 2023 & 0.4329 & 0.5906 & 1.1939 & 0.1298 \\
    MTR v3\cite{shi2024mtra}& 1st 2024 & \textbf{0.4859} & \textbf{0.5554} & \textbf{1.1062} & \textbf{0.1098}\\
    \bottomrule
  \end{tabular}}
\end{table}

\begin{figure}[]
	\centering	
	\includegraphics[scale=0.21]{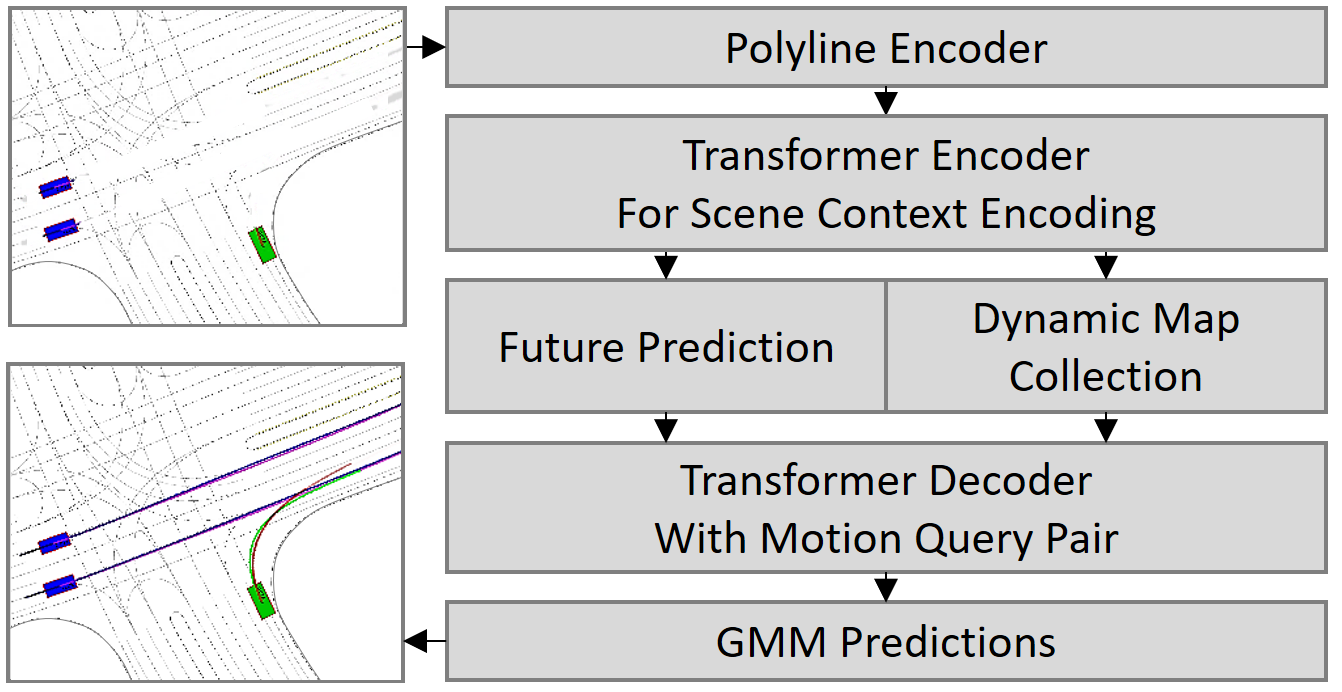}
	\caption{Simplified representation of MTR predictor \cite{ullrich2024transfer}.}
	\label{fig:MTR_arch}
\end{figure}

The temporal-spatial task of trajectory prediction is addressed by MTR through a dedicated transformer design along combining goal-based and direct regression approaches through joint optimization. The overall MTR architecture is illustrated in a simplified manner in Figure \ref{fig:MTR_arch}. The framework leverages vectorized \cite{gao2020vectornet} and agent-centric normalized \cite{zhao2021tnt} polylines $S_i \in \mathbb{R}^{N_i \times P_i \times C_i}$ with $i\in\{a,m\}$ for agents and map, where $N_i, P_i, C_i$ denote the number of agents, map polylines, and state information, respectively. A polyline encoder 
\begin{equation}\label{eq:PolyEncoder_2}
    F_i =  \phi ( \mathrm{MLP} \bigl( S_i \bigr)), \quad i\in\{a,m\},
\end{equation}
based on a multilayer perceptron network $\mathrm{MLP}$ along with max-pooling $\phi$, transforms $S_i$ into $D$-dimensional agent and map features $ F_{i} \in \mathbb{R}^{N_i \times D}$, $i\in\{a,m\}$. For scene context consideration, concatenated features $F_{c} \in \mathbb{R}^{(N_a + N_m) \times D}$ serve as transformer input tokens, that are encoded via $j$-th individual multi-head self-attention \cite{vaswani2017attention} encoder layers $\mathrm{MHSA}(Q_{\mathrm{att}}, K_{\mathrm{att}}, V_{\mathrm{att}})$ with query $Q_{\mathrm{att}}$, key $K_{\mathrm{att}}$, and value $V_{\mathrm{att}}$. Thus, the encoder yielding token features $F'_{c} \in \mathbb{R}^{(N_a + N_m) \times D}$ that are enhanced by context-based future predictions $\mathrm{CTP}_{1:T_{\mathrm{pre}}}$, where $1:T_{\mathrm{pre}}$ denotes the time steps of 2D state-velocity trajectories. Finally, the $\mathrm{CTP}$-trajectories are featurized using (\ref{eq:PolyEncoder_2}), concatenated with $F'_a$, and then processed through an MLP to update the agent tokens $F'_a$, which now include both historical and future trajectories.

Thereafter the generated encodings are decoded along a novel motion query pair of static intention query $Q_{I}$, and dynamic search query $Q_{S}$, which is refined throughout the decoder layers. Thereby, the individual $j$-th decoder layers first apply $Q_{I}$ in multi-head self-attention $\mathrm{MHSA}(C^{j-1} + Q_I, C^{j-1} + Q_I, C^{j-1})$, where $C^{j-1} \in \mathbb{R}^{K_{\mathrm{pre}} \times D}$ denotes query content features, and results in the updated query content $C^j_{qc}$. Second, $Q^{j}_{S}$ is used along the updated query content $C^{j}_{qc}$ and agent tokens $F'_a$ in multi-head cross-attention modules $ \mathrm{MHCA}(Q_{\mathrm{att}}, K_{\mathrm{att},i}, V_{\mathrm{att}})$. Third, the concatenation and MLP processing of $C^{j}_{i}$ yields the query content of the $j$-th layer $C^j \in \mathbb{R}^{K_{\mathrm{pre}} \times D}$ and results in future trajectory predictions after a secondary MLP processing. 

In general, the $K_{\mathrm{pre}}$-modal trajectories are provided by two-dimensional Gaussian components over the prediction horizon $T_{\mathrm{pre}}$, denoted as $\mathcal{N}_{1:K_{\mathrm{pre}}}(\mu_{\mathrm{x}}, \sigma_{\mathrm{x}}; \mu_{\mathrm{y}}, \sigma_{\mathrm{y}}; \rho)$, with means $\mu_{\mathrm{x}}, \mu_{\mathrm{y}}$, standard deviations $\sigma_{\mathrm{x}}, \sigma_{\mathrm{y}},$ and the correlation coefficient $\rho$. In the end, the predicted occurrence probability density of the agents' spatial position at time step $t_{\mathrm{pre}}$ is given by
\begin{equation}\label{eq:Trajectory_MTR}
    \Pr_{t_{\mathrm{pre}}}(\boldsymbol{o}) = \sum_{k_{\mathrm{pre}}=1}^{K_{\mathrm{pre}}} p_{k_{\mathrm{pre}}} \cdot  \mathcal{N}_{k_{\mathrm{pre}}}(o_{\mathrm{x}} - \mu_{\mathrm{x}}, \sigma_{\mathrm{x}}; o_{\mathrm{y}} - \mu_{\mathrm{y}}, \sigma_{\mathrm{y}}; \rho),
\end{equation}
where $\boldsymbol{o}\in\mathbb{R}^2$ denotes the spatial agent position.

While this provides a condensed overview of the MTR’s structure and functionality, we refer the reader to \cite{ullrich2024transfer} for a more detailed summary of MTR, to the original works \cite{shi2022motion, shi2024mtrpp} for full architectural details, and to \cite{huang2023review, barrios2024deep} for a broader discussion on the motion prediction task in general.

\subsection{Model Predictive Path Integral-Based Trajectory Planning}
For local decision-making in dynamic environments, trajectory planning is crucial. For such purposes, the paradigm of model predictive control (MPC) \cite{camacho2007introduction}, that is characterized through horizon-based optimization while directly considering constraints, is well established \cite{guo2018simultaneous, zuo2020mpc, dixit2019trajectory}. Thus, diverse objectives such as safety, efficiency, or comfort could be considered along constraints like steering, and acceleration limits. On this basis, the future state of the vehicle is evaluated in the environment and optimized in a cost-driven manner, i.e. minimized. 

In the application of automated driving (AD), accounting for the nearby evolution of the dynamic environment is essential. Therefore, the multimodal trajectory predictions of other road agents are considered within the costs. As a result, the optimization landscape is highly complex, non-convex, and rapidly changing. For this reason, gradient-free sampling-based approaches \cite{arslan2017sampling, williams2016aggressive, ogretmen2024sampling} such as model predictive path integral control (MPPI) are common in trajectory planning \cite{williams2017model, williams2018information}. While this approach is computationally more demanding than classical gradient-based methods, it allows overcoming local minima, naturally incorporates uncertainties, and is not subject to cost function restrictions. In particular, the consideration of uncertainties is crucial as increasing automation limits or even eliminates human intervention. Therefore, it is important to consider different possibilities and optimize within the imaginable space.

The stochastic optimal control problem (OCP) of MPPI, as introduced by \cite{williams2016aggressive, williams2018information}, is solved over a finite time horizon by iteratively sampling $K_{\mathrm{pla}}$ perturbed input trajectories for a discrete-time dynamical system
\begin{align}
    \boldsymbol{x}_{t_{\mathrm{pla}}+1} = \boldsymbol{f}(\boldsymbol{x}_{t_{\mathrm{pla}}},\boldsymbol{u}_{t_{\mathrm{pla}}} + \boldsymbol{v}_{t_{\mathrm{pla}}}),
    \label{eq:dynamics}
\end{align}
with system states $\boldsymbol{x}_{t_{\mathrm{pla}}} \in \mathbb{R}^{N_{\boldsymbol{x}}}$, control inputs $\! \boldsymbol{u}_{t_{\mathrm{pla}}} \in \mathbb{R}^{N_{\boldsymbol{u}}}$, and random control input perturbations $\boldsymbol{v}_{t_{\mathrm{pla}}}$, drawn from $\mathcal{N}(\boldsymbol{0}, \boldsymbol{\Sigma})$. In the task at hand, a kinematic single-track model is used as the discrete-time dynamical system (\ref{eq:dynamics}), following \cite{althoff2017commonroad}, in combination with a zero-mean Gaussian. 
Consequently, the system states and inputs are given as $ \boldsymbol{u}_{t_{\mathrm{pla}}}=[v_{\delta,t_{\mathrm{pla}}}, a_{\mathrm{long},t_{\mathrm{pla}}}], \boldsymbol{x}_{t_{\mathrm{pla}}}=[x_{t_{\mathrm{pla}}}, y_{t_{\mathrm{pla}}}, \delta_{t_{\mathrm{pla}}}, v_{t_{\mathrm{pla}}}, \Psi_{t_{\mathrm{pla}}}]$. For trajectory planning, the optimal control input sequence $\boldsymbol{U}^* \in \mathbb{R}^{N_\mathrm{\boldsymbol{u}}\times  T_{\mathrm{pla}}}$ is determined by
\begin{align}
    \boldsymbol{U}^*=\underset{\boldsymbol{U} \in {\mathcal{U}}}{\text{argmin}} \,\, \mathbb{E}_{\mathbb{Q}}\bigl[ \underbrace{\textstyle\sum_{t_{\mathrm{pla}}=0}^{T_{\mathrm{pla}}-1} l(\boldsymbol{x}_{t_{\mathrm{pla}}},\boldsymbol{u}_{t_{\mathrm{pla}}}}_{S(\boldsymbol{X},\boldsymbol{U})})\bigr],
    \label{eq:MPC_problem}
\end{align}
where $l(\boldsymbol{x}_{t_{\mathrm{pla}}}, \boldsymbol{u}_{t_{\mathrm{pla}}})$ denote the running costs, $\mathcal{U}$ the set of admissible control input sequences, and $T_{\mathrm{pla}}$ the planning horizon. The application-specific constraints are  considered according to \cite{althoff2017commonroad}. Based on the total trajectory cost $S(\boldsymbol{X},\boldsymbol{U})$, each of the $K_{\mathrm{pla}}$ trajectory rollouts is assigned a weighting factor given by
\begin{align}
    w^{(k_{\mathrm{pla}})} = \text{exp}(-\frac{1}{\lambda_{\mathrm{iT}}}S^{(k_{\mathrm{pla}})})), 
    \label{eq:weight_determination}
\end{align}
with the inverse temperature $\lambda_{\mathrm{iT}} \in \mathbb{R}^+$, and $k_{\mathrm{pla}}=\{1,\dots, K_{\mathrm{pla}}\}$. The optimal input trajectories are derived by an iterative averaging scheme update
\begin{align}
    \boldsymbol{u}_{t_{\mathrm{pla}}}^* \gets \boldsymbol{u}_{t_{\mathrm{pla}}}^* + \frac{\sum_{k_{\mathrm{pla}}=1}^{K_{\mathrm{pla}}} w^{(k_{\mathrm{pla}})}{\boldsymbol{v}_{t_{\mathrm{pla}}}^{(k_{\mathrm{pla}})}}}{\sum_{k_{\mathrm{pla}}=1}^{K_{\mathrm{pla}}} w^{(k_{\mathrm{pla}})}}. 
\end{align}
Across time steps, inspired by \cite{kingma2014adam}, a momentum-based update is used with the factor $\beta_{\mathrm{pla}}$. Overall, the optimal input trajectory determination in MPPI is based on minimizing the KL divergence $D_{\text{KL}}(\mathbb{Q}||\mathbb{Q}^*)$, where $\mathbb{Q}^*$ represents the latent optimal distribution \cite{williams2018information}. The theoretical background is further detailed in \cite{williams2016aggressive, williams2018information}.

\subsection{Integration \& Implementation}
Since the transition from open-loop to closed-loop has a significant impact on the performance of trajectory prediction and thus planning \cite{dauner2023parting, caesar2021nuplan}, we consider the more challenging but also more realistic use case of closed-loop simulation. For this purpose, the initial MTR baseline model was transfer-learned according to \cite{ullrich2024transfer}, considering the best performing fine-tuned model in this study. Furthermore, the MPPI planning is implemented as outlined previously. The integration of the MTR predictions into the MPPI planning is achieved via the cost function 
\begin{align}\label{eq:costfunction_MPPI}
	\begin{split}
    l(\boldsymbol{x}_{t_{\mathrm{pla}}},\boldsymbol{u}_{t_{\mathrm{pla}}}) &=  l_{t_{\mathrm{pla}}}^{\mathrm{safe}} + l_{t_{\mathrm{pla}}}^{\mathrm{prog}} + l_{t_{\mathrm{pla}}}^{\mathrm{comf}} + l_{t_{\mathrm{pla}}}^{\mathrm{norm}}, \\
    l_{t_{\mathrm{pla}}}^{\mathrm{safe}}&= e^{-t_{\mathrm{pla}}\cdot\lambda_d} \cdot (l_{b,t_{\mathrm{pla}}} + l_{\Pi,t_{\mathrm{pla}}} + l_{\Psi, t_{\mathrm{pla}}}),\\
    l_{t_{\mathrm{pla}}}^{\mathrm{prog}}&= e^{-t_{\mathrm{pla}}\cdot\lambda_d} \cdot (l_{p,t_{\mathrm{pla}}}),\\
    l_{t_{\mathrm{pla}}}^{\mathrm{comf}}&= e^{-t_{\mathrm{pla}}\cdot\lambda_d} \cdot (l_{c,t_{\mathrm{pla}}} + l_{i, t_{\mathrm{pla}}}),\\
    l_{t_{\mathrm{pla}}}^{\mathrm{norm}}&= e^{-t_{\mathrm{pla}}\cdot\lambda_d} \cdot (l_{o,t_{\mathrm{pla}}} + l_{v,t_{\mathrm{pla}}}),
	\end{split}
\end{align}
where safety, progress, comfort, and normative behavior costs comprise several distinct sub-costs, which are explained in the following. But first, it should be noted, that all sub-costs are decayed over the trajectory horizon depending on the factor $\lambda_d$. Additionally, several sub-costs incorporate penalty functions, which are described before addressing the sub-costs in detail. The boundary and closeness penalties are given by
\begin{align}\label{eq:penalty}
	\begin{split}
   P_{\mathrm{bnd}}(\text{arg}) &= \hat{p} \left( \mathrm{sig}(\hat{s} \cdot\text{arg}) + \log(1 +\hat{s} \cdot \overline{\exp}(\text{arg} - \hat{h})) \right),\\
   P_{\mathrm{cls}}(\text{arg}) &= \frac{1}{1+(\text{arg})^2} 
	\end{split}
\end{align}
where $\mathrm{sig}= 1/(1+\exp(-x))$ denotes the sigmoid function, and the parameters $\hat{p}=100, \hat{s}=25, \hat{h}=7.5$, represent the penalty, scale, and shift, respectively. Additionally, $\overline{\exp}(\text{arg})=\exp(\min(\text{arg},700))$ defines the upper-bounded exponential function with a limit of  $700$.

The safety costs encompass boundary $l_{b,t_{\mathrm{pla}}}$, traffic $l_{\Pi,t_{\mathrm{pla}}}$, and yaw $l_{\Psi, t_{\mathrm{pla}}}$ cost terms, stated as
\begin{align}\label{eq:costs_safety_costs}
	\begin{split}
    l_{b, t_{\mathrm{pla}}} &= \xi_{b} \cdot \textstyle\sum_{p \in \{\mathrm{bnd}, \mathrm{cls}\}} ( P_{p}(\Delta \underline{b}_{t_{\mathrm{pla}}}) + P_{p}(\Delta \overline{b}_{t_{\mathrm{pla}}}))\\
    l_{\Pi, t_{\mathrm{pla}}}&=  \xi_{\Pi} \cdot \textstyle\sum_{a=1}^{N_{a}}\textstyle\sum_{k_{\mathrm{pre}}=1}^{K_{\mathrm{pre}}} c_{a,k_{\mathrm{pre}}} \cdot \\ & \quad \left( P_{\mathrm{bnd}}(-d_{t_{\mathrm{pla}}}^{e|a,k_{\mathrm{pre}}}) + P_{\mathrm{cls}}((\overline{d}^{e|a}-d_{t_{\mathrm{pla}}}^{e|a,k_{\mathrm{pre}}})\right)\\
    l_{\Psi, t_{\mathrm{pla}}} &= \xi_{\Psi} \cdot (\Delta \Psi_{t_{\mathrm{pla}}})^2,
	\end{split}
\end{align}
with the respective tunable weights $\boldsymbol{\Xi}_{\mathrm{safe}}=\{\xi_b, \xi_{\Pi}, \xi_{\Psi}\}$. Boundary costs $l_{b,t_{\mathrm{pla}}}$ are derived from penalty functions (\ref{eq:penalty}) based on the left boundary distance $\Delta\underline{b}_{t_{\mathrm{pla}}}=\underline{d}_{\mathrm{lat}} - {d}_{\mathrm{lat},t_{\mathrm{pla}}}$, and right boundary distance $\Delta\overline{b}_{t_{\mathrm{pla}}}={d}_{\mathrm{lat},t_{\mathrm{pla}}} - \overline{d}_{\mathrm{lat}}$. Here, $d_{\mathrm{lat},t_{\mathrm{pla}}},\underline{d}_{\mathrm{lat}}, \overline{d}_{\mathrm{lat}}$ denote the lateral offset from a given route at time $t_{\mathrm{pla}}$, the lateral left boundary, and the lateral right boundary.

The traffic costs $l_{\Pi,t_{\mathrm{pla}}}$ integrate MTR trajectory predictions $\Gamma_{\mathrm{pre}}$, generated at time step $t_{\mathrm{pla}}=0$, for all agents $N_{a}$ within an attention radius of $\SI{50}{\metre}$. For each agent $a$, costs are calculated for each modality $k_{\mathrm{pre}}$, which are weighted by the associated confidence $c_{a,k_{\mathrm{pre}}}$. The costs build upon boundary and closeness penalties from (\ref{eq:costs_safety}). In doing so, the distance between the ego vehicle and the respective agent is given by
\begin{align}\label{eq:delta_d}
	\begin{split}
d_{t_{\mathrm{pla}}}^{e|a,k_{\mathrm{pre}}} &= \min_{\boldsymbol{p}_1 \in\mathcal{B}^{e}, \: \boldsymbol{p}_2 \in\mathcal{B}^{a,k_{\mathrm{pre}}}}\| \boldsymbol{p}_{1}(t_{\mathrm{pla}}) - \boldsymbol{p}_{2}(t_{\mathrm{pla}}) \|,
	\end{split}
\end{align}
where $\mathcal{B}^{e}\in\mathbb{R}^2$ denotes the ego bounding box set depending on the augmented state vector $\boldsymbol{\tilde{x}}^{e}_{\{\mathrm{x,y,\Psi},w,l,s\}}(t_{\mathrm{pla}})$, that also considers the ego width $w$, length $l$, and additional longitudinal safety distance $s=\SI{0.125}{\metre}$. Respectively, $\mathcal{B}^{a,k_{\mathrm{pre}}}\in\mathbb{R}^2$ denotes the agent's boundary set that is defined in accordance to the predicted trajectory state $ \boldsymbol{\tau}_{\{\mathrm{x,y,\Psi,w,l}\}}^{a,k_{\mathrm{pre}}}(t_{\mathrm{pre}}=t_{\mathrm{pla}})$.
In accordance, the distance threshold between the ego and agents is denoted as $\overline{d}^{e|a}$. Moreover, the yaw costs $l_{\Psi, t_{\mathrm{pla}}}$ are based on the squared yaw deviation $\Delta \Psi_t =\mathrm{wrap}( {\Psi}_{t} - {\Psi}_{\mathrm{ref},t})$. Here, ${\Psi}_{\mathrm{ref},t}$ denotes the orientation of the respective route segment.

The progress costs are derived through
\begin{align}\label{eq:costs_safety}
	\begin{split}
    l_{p,t_{\mathrm{pla}}}&= \xi_{p} \cdot (1 - \frac{l_{w,t_{\mathrm{pla}}}}{L_{w}}),
	\end{split}
\end{align}
with $l_{w,t_{\mathrm{pla}}}$ denoting the route index of the trajectory at time step $t_{\mathrm{pla}}$ given by the projecting on the route, and $L_w$ the index of the endpoint within a moving local window. Here, $\boldsymbol{\Xi}_{\mathrm{prog}}=\{\xi_p\}$, denotes the tunable progress parameter.

For comfort, the costs $l_{c,t_{\mathrm{pla}}}, l_{i,t_{\mathrm{pla}}}$ are aimed at smoothing both input and output according to
\begin{align}\label{eq:costs_safety}
	\begin{split}
    l_{c,t_{\mathrm{pla}}}&= \xi_{c} \cdot ({d}_{\mathrm{lat},t_{\mathrm{pla}}} - {d}_{\mathrm{lat},t_{\mathrm{pla}}-1})^2,\\
    l_{i,t_{\mathrm{pla}}}&= \xi_{i}  \cdot \|\boldsymbol{u}_{t_{\mathrm{pla}}} - \boldsymbol{u}_{t_{\mathrm{pla}}-1}\|^2_2,
	\end{split}
\end{align}
with the tunable parameters $\boldsymbol{\Xi}_{\mathrm{comf}}=\{\xi_i, \xi_c\}$.

Normative behavior costs include lateral offset costs $l_{o,t_{\mathrm{pla}}}$, and curvature-aware velocity costs $l_{v,t_{\mathrm{pla}}}$, expressed as
\begin{align}\label{eq:costs_safety}
	\begin{split}
    l_{o,t_{\mathrm{pla}}}&= \xi_{o} \cdot (d_{\mathrm{lat},t_{\mathrm{pla}}})^2,\\
    l_{v,t_{\mathrm{pla}}} &= \xi_{v} \cdot \left((v_{t_{\mathrm{pla}}} - v_{\mathrm{des}})^2 + P_{\mathrm{bnd}}(-v_{t_{\mathrm{pla}}})\right),
	\end{split}
\end{align}
with the tunable parameters $\boldsymbol{\Xi}_{norm}=\{\xi_o, \xi_v\}$. Offset costs penalize lateral deviations $d_{\mathrm{lat}}$, while velocity costs are based on deviations from the desired velocity $v_{\mathrm{des}}$. The desired velocity is determined by the permissible speed, taking into account a limitation on lateral acceleration of $\SI{4}{\meter\per\second\squared}$, thereby incorporating curvature-awareness.
\section{TrustMHE}\label{methodology}
In research, data-driven statistical AI systems for motion prediction and their integration into stochastic optimization-driven planners are widely used. The deployment of such approaches, particularly in high-risk applications, requires a high level of trust in their functionality. One way to address these concerns is by enhancing self-awareness and trust in AI. To this end, the TrustMHE method is introduced in this section. First, some core principles are discussed, followed by an explanation of the methodological concept, and concluding with an overview of the implementation.

\subsection{Core Principles}
The automation of safety-critical cyber-physical systems increasingly takes uncertainties into account\cite{van2023autotech}. To avoid misconceptions, it is essential to supplement the existing payload data with quality data \cite{van2023autotech}, e.g. a measure of credibility. In the given use case, quality data enables the identification of MTR performance degradation caused by distribution shifts. By integrating self-assessment-driven quality data, the system self-awareness can be enhanced, enabling timely interventions to mitigate potential damage and ultimately strengthening trust in AI.

For this purpose, an online reliability estimation method based on Subjective Logic (SL) \cite{josang2016subjective} has recently been proposed by \cite{josang2016subjective,muller2019subjective}. SL extends classical probability theory by incorporating statistical uncertainties and subjective beliefs, making it well-suited for processing incomplete, unreliable, or subjective information. While the SL method is particularly suitable for the fusion of different beliefs \cite{muller2019subjective, griebel2020kalman}, i.e. subjective self-assessments, an online reliability assessment for individual AI systems can be carried out according to this way of thinking. Such reliability uncertainty can be natively integrated into the overall system (self-assessment and self-awareness).

The current methodology-agnostic self-assessment of AI systems primarily relies on OOD detection and safety monitors. These techniques focus on detecting deviations in payload data from training data distributions by monitoring input data \cite{liu2020input, sabokrou2018adversarially}, and/or internal states \cite{cheng2019runtime, henzinger2019outside}, and/or output data \cite{hendrycks2016baseline, liang2017enhancing}, depending on the approach. While these methods have gained some acceptance, they do not offer a systematic approach to handling degradation. However, in the field of control and safety engineering, threshold-based switching to a fallback path has been proposed \cite{schoning2023safe}.

Within safety-critical cyber-physical systems, particularly in closed-loop operation, control engineering serves to ensure safety, reliability, and trustworthiness. The moving horizon estimation (MHE) is a method that uses a sliding window over a past horizon to continuously compare the output of a dynamic system model with measurements and solve an optimization problem to refine the model by estimating system states and parameters \cite{findeisen1997moving, allan2019moving}. While this approach systematically counteracts degradation through state and parameter estimation and actualization, it is not directly applicable to AI systems. However, it provides a solid methodological basis for online reliability estimation in real world closed-loop applications.

It turns out that different core principles exist, each offering unique advantages, but they have not yet been merged adequately. The methodical fusion of the core principles and thus the concept of the TrustMHE is described in the following.

\subsection{Methodological Concept}
One of the key challenges of AI systems is that they are typically exposed to dynamic environments, where inputs and conditions cannot be controlled to remain within the training distributions where assumptions hold. Moreover, AI systems cannot be quickly adjusted to realign assumptions, making the problem even more challenging. As a result, the classical MHE approach, which minimizes the mismatch between measurements and model outputs by estimating current states and parameters, is not directly applicable. Nevertheless, when estimating the reliability of an AI and considering the necessity of a fallback, as outlined in \cite{schoning2023safe}, the reliability estimate could directly serve as a model blending parameter. Consequently, the overall module can be conceptualized as
\begin{align}\label{eq:new_model}
	\begin{split}
    f_{\mathrm{TrustAI}} &=  \omega \cdot f_{\mathrm{AI}} + (1-\omega) \cdot f_{\neg\mathrm{AI}},\\
	\end{split}
\end{align}
with the probability-based reliability estimate $\omega \in [0,1]$, which combines the AI-based system $f_{\mathrm{AI}}$ with an AI-free fallback system $f_{\neg\mathrm{AI}}$. As a result and in addition to facilitating an adjustable overall model, the degradation handling is systematically incorporated by adhering to a probability-based decision-making design, which enables seamless overall integration. The determination of the reliability estimate is given by
\begin{align}\label{eq:disc}
	\begin{split}
    \omega &= \phi\Bigl( \textstyle\sum_{t_{\mathrm{est}}=t-T_{\mathrm{est}}}^{t} D\bigl(f_{\mathrm{TrustAI}}(t_{\mathrm{est}}),  f_{\mathrm{meas}}(t_{\mathrm{est}})\bigr)\Bigr) ,\\
	\end{split}
\end{align}
where $\phi: \mathbb{R}^+ \to [0, 1]$ denotes a mapping from a non-negative scalar discrepancy measure, derived from the discrepancy function $D$, to the probability-based reliability estimate $\omega$. Thereby, equation (\ref{eq:disc}) facilitates to fuse different concepts illustrated in the previous subsection. For example, if the discrepancy function $D$ is considered as a general cost function that quantifies the error or mismatch between predicted and observed states over a past time horizon, and $\phi$ is interpreted as a minimization operator with constraints (e.g., $\min_{\omega}, \mathrm{s.t.}\: 0\leq\omega\leq1$), the framework closely resembles classical MHE. In contrast, by using a discrepancy function like the Kullback-Leibler (KL) divergence, $D_{\mathrm{KL}}$, in combination with an activation function like the sigmoid for $\phi$, the framework allows for distribution-based estimation (e.g., $D_{\mathrm{KL}}(f_{\mathrm{TrustAI}},  f_{\mathrm{meas}}|\omega=1)$). This approach aligns more closely with techniques like OOD detection and safety monitors, particularly when the moving horizon is reduced to a single time step. 

Thus, while relaxing the requirement of optimization-based estimation, TrustMHE still retains many characteristics of classical MHE, while simultaneously merging core principles and fostering broader application. The following use case demonstrates that the generic abstraction, along with the resulting flexibility, is crucial for ensuring applicability across various AI systems.

\subsection{Implementation \& Integration}\label{TrustMHEImplementationIntegration}
Within the given use case, the concept of TrustMHE should be applied to the AI-based MTR prediction model \cite{shi2022motion, ullrich2024transfer} in order to increase system self-awareness and trust in AI. In doing so, there are a variety of use case-specific design decisions:
\begin{enumerate}[label=\textbf{Q\arabic*},  left=0.5em]
  \item \label{q:1}What should be used as a fallback model?
  \item \label{q:2}How should the discrepancy function be defined?
  \item \label{q:3}How is an uncertainty-aware multimodal prediction compared with unimodal measurements?
  \item \label{q:4}What should be considered within the moving horizon?
\end{enumerate}

Regarding (\ref{q:1}), the simplest fallback model would consider the road users as static, making no predictions, increasing the safety distances, and ensuring no crashes. The result would be an extremely conservative driving behavior, possibly even a robot freezing problem \cite{trautman2010unfreezing}. This would not only lead to traffic obstructions but also result in unpredictable ego behavior for other road users, creating an unsafe overall situation. Such an approach would therefore be neither practical nor a fair basis for comparison. Instead, we employ a generally recognized physical model, the constant velocity model, which provides reliable results over short time horizons by assuming constant longitudinal and steering angle velocities.

For the discrepancy function (\ref{q:2}), various approaches are conceivable. However, due to unimodality of fallback predictions and measurements in contrast to the uncertainty-aware multimodalities of the AI-based MTR predictions, a direct optimization-based approach is not suitable. Neither is a classical distribution-based approach appropriate, as today's AI-based prediction models are aimed at generalization, and therefore a deviation from the training distribution does not necessarily lead to a performance drop.

An intuitive approach is the integrating of existing prediction evaluation metrics into an online reliability estimation. While a binary classification, such as miss Rate (MR) \cite{ettinger2021large}, is conceivable, displacement errors, e.g. minADE and minFDE \cite{ettinger2021large}, are better suited for a continuous reliability measure. Although the minimum displacement error, i.e., the minimum over the modalities $k_{\mathrm{pre}}$, is a widely used metric, such a metric neglects the fact that the modalities are weighted by a confidence $c_{k_{\mathrm{pre}}}$ in the MPPI problem's cost function (\ref{eq:costfunction_MPPI}). Thus, with regard to (\ref{q:3}), the displacement error of each modality, weighted by the respective confidence, should be taken into account. Furthermore, according to the moving horizon concept, rather than using a discrete or final displacement error, an average over the TrustMHE horizon $T_{\mathrm{est}}$ should be used. Accordingly, a weighted average displacement error is calculated as
\begin{align}\label{eq:JJ}
	\begin{split}
    d_{\mathrm{t{'}}}^{N_{\tilde{a}}} &= \frac{1}{N_{\tilde{a}} \cdot K_{\mathrm{pre}}} \sum_{\tilde{a}=1}^{N_{\tilde{a}}} \sum_{k_{\mathrm{pre}}=1}^{K_{\mathrm{pre}}}\frac{c_{\tilde{a},k_{\mathrm{pre}}}}{T_{\mathrm{est}}} \sum_{t_{\mathrm{est}}=t{'}-T_{\mathrm{est}}}^{t{'}} d_{t_{\mathrm{est}}}^{\hat{a}|\tilde{a},k_{\mathrm{pre}}},
	\end{split}
\end{align}
where $t{'}$ denotes the current time step, represented in the planning application as $t_{\mathrm{pla}}=0$. Measured agent trajectories are represented as $\hat{a}$, while trajectory predictions generated at time step $t'-T_{\mathrm{est}}$ are denoted as $\tilde{a}$. The respective total number of predicted agents is given by $N_{\tilde{a}}$. The displacement error $d_{t_{\mathrm{est}}}^{\hat{a}|\tilde{a},k_{\mathrm{pre}}}$ is computed for each measured agent $\hat{a}$ by comparing it to all corresponding modalities $\tilde{a},k_{\mathrm{pre}}$ at each time step $t_{\mathrm{est}}$ along the trajectory up to the current time step $t'$. The cumulative displacement error over the estimation horizon for each agent's predicted modality is weighted by the respective confidence, denoted as $c_{\tilde{a},k_{\mathrm{pre}}}$.

With regard to (\ref{q:4}), the predictions considered in (\ref{eq:JJ}) are generated by MTR at time step $t'-T_{\mathrm{est}}$, while applying the moving horizon across the entire trajectory. Contrary to this would be the application of the moving horizon over multiple MTR prediction steps, while considering only a single displacement error per trajectory at a discrete point in time. Although in theory the moving horizon could be applied to both, practically due to the computational effort, this is not applicable. Therefore, as greater importance is assigned to the trajectory, the first approach is considered. Furthermore, to account also for the effects across the MTR prediction steps, a momentum-inspired \cite{kingma2014adam} update of the resulting reliability estimate is performed, given by 
\begin{align}\label{eq:costfunction_MPPI}
	\begin{split}
     \omega_{t'_{\mathrm{pla}}} &=  \beta_{\mathrm{est}} \cdot \omega_{t'_{\mathrm{pla}}-1} + (1-\beta_{\mathrm{est}}) \cdot \gamma(-d_{\mathrm{t{'}}}^{N_{\tilde{a}}}),\\
	\end{split}
\end{align}
where the mapping function is defined as $\gamma:=2 \, \mathrm{sig}(\cdot)$, and the momentum factor is given by $\beta_{\mathrm{est}}$.

To sum up, it can be seen, that due to the handling of different representations of measurements and predictions, the displacement function operates within a transferred problem space, namely the traffic costs. Since the weighted averaged displacement error calculation is similar to the relevant MPPI optimization problem, the MTR-specific TrustMHE implementation implicitly exhibits optimization-based characteristics. However, as w.r.t. MTR prediction, a single time step prevails, alongside an AI-centric evaluation, the implementation also reflects inherently the characteristics of OOD detection and safety monitoring. Thus, the task-specific TrustMHE highlights both the need for flexibility and the potential of the outlined concept. This emphasizes not only the task-specific requirement for methodological adaptability but also the concept's ability to integrate and harmonize different core principles.

Finally, it should be noted, that the TrustMHE integration occurs within the transformed problem space, updating the traffic costs in the MPPI problem (\ref{eq:costs_safety_costs}) to:
\begin{align}\label{eq:costfunction_MPPI_update}
	\begin{split}
    l_{\Pi,t_{\mathrm{pla}}}^{\mathrm{TrustMHE}} &=  \omega_{t'_{\mathrm{pla}}} \cdot l_{\Pi, t_{\mathrm{pla}}} + (1-\omega{_{t'_{\mathrm{pla}}}}) \cdot 
    l_{\zeta, t_{\mathrm{pla}}},\\
	\end{split}
\end{align}
with $l_{\zeta,t_{\mathrm{pla}}}$ denoting the costs given by the fallback model.
\section{Empirical Evaluation}\label{analysis}
In this section, the experimental setup is outlined, along with the presentation of results.

\subsection{Experimental Setup}
\begin{figure}[]
	\centering	
    \resizebox{0.46\textwidth}{!}{
        \begin{tikzpicture}[
node distance=5cm,
box/.style={
    draw,
    rectangle,
    minimum width=2.5cm,
    minimum height=1.5cm,
    align=center,
    rounded corners
},
arrow/.style={
    ->,
    thick,
    >=latex
}
]

\node[box] (carmaker) {CarMaker\\Simulator};
\node[box, right of=carmaker] (planner) {MPPI\\Planner};
\node[box, right of=planner] (mtr) {MTR\\Model};
\node[box, above of=carmaker, node distance=2.5cm] (eval) {Evaluation\\Script};

\node[box, below of=mtr, node distance=2.5cm] (mtr_cost) {MTR\\Traffic Cost};
\node[box, below of=planner, node distance=2.5cm] (mhe) {MHE Adjusted Cost};
\node[box, below of=mtr_cost, node distance=1.5cm] (cv_cost) {CV\\Traffic Cost};

\draw[arrow] ([yshift=-0.2cm]carmaker.north east) -- node[above] {State} ([yshift=-0.2cm]planner.north west);
\draw[arrow] ([yshift=0.2cm]planner.south west) -- node[below] {Trajectory} ([yshift=0.2cm]carmaker.south east);

\draw[arrow] (planner) -- node[above] {State Update} (mtr);
\draw[arrow] (mtr) -- node[right] {Predictions} (mtr_cost);

\draw[arrow] (mtr_cost) -- node[above] {Cost} (mhe);
\draw[arrow] (cv_cost) -- ++(-5, 0) -- node[left] {Cost} (mhe.south);
\draw[arrow] (mhe.north)  -- node[right] {Cost} (planner.south);

\draw[arrow] (eval.south) -- node[left] {Monitor} (carmaker);
\draw[arrow] (eval.east) -- ++(3.75, 0) -- node[above right] {Monitor} (planner);

\node[text width=3cm, align=center] at ($(carmaker)!0.5!(planner)$) {\small TCP Comm.};

\node[text width=5cm, align=left] at ($(carmaker.south) + (1.25cm,-2cm)$) {\small{
    Update intervals:\\
    \begin{itemize}
    \item State - \textbf{5 ms}
    \item Trajectory - \textbf{100 ms}
    \item Prediction - \textbf{250 ms}
    \end{itemize}
}};

\end{tikzpicture}
    }
	\caption{Illustration of the closed-loop experimental setup.}
	\label{fig:Eval}
\end{figure}
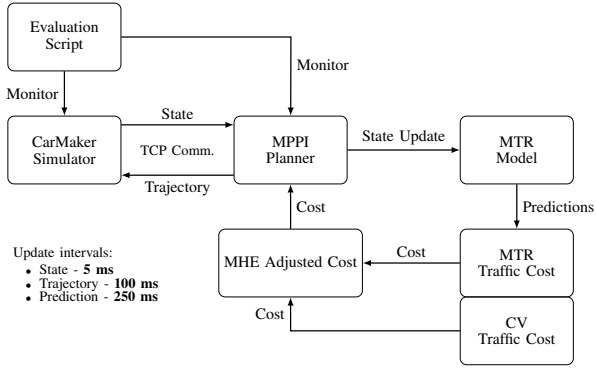

An overview of the experimental setup for the trajectory prediction and planning case study is shown in Figure \ref{fig:Eval}. The TrustMHE approach is evaluated in a closed-loop simulation, as prior studies \cite{dauner2023parting, caesar2021nuplan} have shown significant differences between open-loop and closed-loop performance. To ensure a realistic assessment, the experimental setup consists of a closed-loop co-simulation based on the CarMaker\footnote{\url{https://www.ipg-automotive.com/en/products-solutions/software/carmaker/}} simulation tool, along with the transfer-learned, pre-trained MTR baseline model \cite{ullrich2024transfer}. Furthermore, to simulate real-world conditions more accurately, the planned trajectory is not executed directly, but instead transmitted to the IPGDriver Trajectory Extension\footnote{\url{https://www.ipg-automotive.com/en/press/carmaker-supports-unicaragil/}} module. The evaluation of the TrustMHE approach, implemented according to Subsection \ref{TrustMHEImplementationIntegration}, focuses on its impact on driving behavior in dynamic environments. The corresponding hypotheses are formulated as follows:
\begin{enumerate}[label=$\boldsymbol{H_\arabic*}$,  left=0.5em, start=0]
  \item \label{h:0}TrustMHE does not improve safety in dynamic environments.
  \item \label{h:1}TrustMHE improves safety in dynamic environments.
\end{enumerate}

In order to ensure valid evaluations and minimize system-specific impacts as much as possible, the TrustMHE approach is evaluated under various conditions. This includes the consideration of realistic traffic (\textit{Stochastic traffic option}) and agent driver behavior (\textit{Human driver behavior}) in CarMaker as well as varying scenarios in terms of topology and velocities. Additionally, to minimize the impact of the planner setup, several planner modes and sampling distributions are considered. A more detailed overview of the setup configuration and setup variations is shown in Table \ref{tab:setup_2}. Moreover, the mode-dependent MPPI cost function parametrization is presented in Table \ref{tab:setup_2}. 

For the evaluation of the hypotheses \ref{h:0} and \ref{h:1}, crashes and the minimum distance to other agents are evaluated. In order to obtain many challenging interactions, the simulation is not terminated in the event of a crash, but continues normally. This approach enables the generation of a larger number of difficult interactions within a single test run, facilitating the use of the crash count, \textsc{Crashes}, as a metric to allow richer comparisons beyond a binary result. Nevertheless, the aggregate binary metric \textsc{Success}, defined as a test run without crashes, is also taken into account. The metric \textsc{Min. Dist.} (minimum distance) is then calculated across all successful test runs. Finally, to evaluate the impact on efficiency, the \textsc{Progress} metric, defined as the relative progress in a time-limited test run, is also considered. This is relevant to determine whether the approach preserves efficiency, as safety improvements shouldn't hinder progress. While separate hypotheses could be formulated for these additional investigations, they are omitted due to space constraints and to maintain clarity by focusing on the key aspect of safety.

\begin{table}[h]
  \centering
  \caption{Case study evaluation setup parametrization.}
  \resizebox{\columnwidth}{!}{
  \label{tab:setup}
  \begin{tabular}{l l l l l}
    \toprule
    \multirow{7}{*}{\rotatebox{90}{Sim.}} & Human driver behavior & & Activated \\
    & Stochastic traffic option& & Activated \\
     & Scenario & & \{Overtake., Junct., Urb.\}&  \\
     & Scenario duration & & \{100, 100, 35\} & $[\SI{}{\second}]$ \\
     & Number of traffic agents & & \{23, 2, 57\} & $[\SI{}{\#}]$  \\
     & Velocities & $v$&  $4 \dots 30$& $[\SI{}{\meter\per\second}]$  \\
     & Road boundaries & $\overline{\underline{d}}_{\mathrm{lat}}$& Scenario-specific & \\
    \midrule
    \multirow{2}{*}{\rotatebox{90}{Pre.}} & MTR Prediction horizon &$T_{\mathrm{pre}}$ & 50 & $[\text{steps}]$ \\
    & Modalities &$K_{\mathrm{pre}}$ & 6 & $[\SI{}{\#}]$ \\
    \midrule
    \multirow{6}{*}{\rotatebox{90}{Pla.}} & MPPI Planning horion& $T_{\mathrm{pla}}$  & 50 & $[\text{steps}]$ \\
    & Rollouts &$K_{\mathrm{pla}}$ & 200 & $[\SI{}{\#}]$ \\
    & Inverse temperature&$\lambda_{\mathrm{iT}}$ & 0.02 \\
    & Weight decay&$\lambda_d$ & 0.05 \\
    & Control momentum&$\beta_{\mathrm{pla}}$ & 0.75 \\
    & Sampling noise&$\sigma_{\mathrm{pla}}$ & \{0.1,1.0\} \\
    \midrule
    \multirow{2}{*}{\rotatebox{90}{Est.}} & TrustMHE horizon &$T_{\mathrm{est}}$ & \{1,3,5,15,30\}& $[\text{steps}]$ \\
    & TrustMHE momentum &$\beta_{\mathrm{est}}$ & 0.25  &\\
    \bottomrule
  \end{tabular}}
\end{table}

\begin{table}[]
  \vspace{2mm}
  \centering
  \caption{Tunable parameters $\boldsymbol{\Xi}$ of MPPI's planner cost function (\ref{eq:costfunction_MPPI}) for different planner modes.}
  \resizebox{\columnwidth}{!}{
  \label{tab:setup_2}
  \begin{tabular}{l|c c c c| c |c  c|c c c }
    \toprule
    \multirow{2}{*}{\diagbox[width=2cm, height=0.65cm, dir=SE]{\textbf{Mode}}{\textbf{Param.}}} & \multicolumn{4}{c|}{$\boldsymbol{\Xi}_{\mathrm{safe}}$}  & \multicolumn{1}{c|}{$\boldsymbol{\Xi}_{\mathrm{prog}}$}& \multicolumn{2}{c|}{$\boldsymbol{\Xi}_{\mathrm{comf}}$} & \multicolumn{3}{c}{$\boldsymbol{\Xi}_{\mathrm{norm}}$} \\
    & $\xi_b$ & $\xi_{\Pi}$ & $\xi_{\Psi}$ & $\xi_{\dot{\Psi}}$ & $\xi_p$ & $\xi_i$ & $\xi_c$ & $\xi_{\Sigma\Psi}$ &  $\xi_o$ & $\xi_v$ \\
    \midrule
    Conservative & 2 & 2 & 1 & 1 & 1 & 1 & 1 & 1 & 1 & 1   \\
    Balanced & 2 & 2 & 1 & 1 & 2 & 1 & 1 & 1 & 1 & 2 \\
    Aggressive & 1 & 1 & 1 & 1 & 2 & 1 & 1 & 1 & 1 & 2 \\
    \bottomrule
  \end{tabular}}
\end{table}

\subsection{Results}

\begin{table*}[ht!]
  \centering
  \caption{Performance metrics comparison.}
  \label{tab:performance_metrics_TrustMHE}
  \begin{tabular}{l r r r r r r r l r r }
    \toprule
    Metric & TrustMHE & runs & mean & median & std & min & max & statistical hypothesis test & statistic & p-value \\
    \midrule
    \multirow{2}{*}{Crash $[\SI{}{\#}]$}& Disabled & 144 & 1.74 & 1.00 & 2.42 & 0.00 & 14.00&  \multirow{2}{*}{Mann–Whitney ${\displaystyle U}$ test}& \multirow{2}{*}{41216.00} & \multirow{2}{*}{0.000028} \\
    & Enabled & 720 & 1.06 & 0.00 & 1.79 & 0.00 & 12.00 & & & \\
    \midrule
    \multirow{2}{*}{Progress $[\SI{}{\percent}]$}& Disabled & 144 & 72.56 & 77.14 & 30.47 & 6.27 & 100.00&  \multirow{2}{*}{Mann–Whitney ${\displaystyle U}$ test}& \multirow{2}{*}{52132.00} &
    \multirow{2}{*}{0.910229} \\
    & Enabled& 720 & 72.96 & 85.43 & 32.52 & 6.02 & 100.00& & & \\
    \midrule
    \multirow{2}{*}{Success $[\SI{}{\percent}]$}& Disabled & 144 & 36.11 & 0.00 & 48.20 & 0.00 & 100.00 &  \multirow{2}{*}{$\chi^2$ test (with Yates’s correction)}& \multirow{2}{*}{9.45} & \multirow{2}{*}{0.002102} \\
    & Enabled & 720 & 50.56 & 100.00 & 50.03 & 0.00 & 100.00& & & \\
    \midrule
    \multirow{2}{*}{Min. Dist. $[\SI{}{\centi\meter}]$ }& Disabled & 52 & 0.82 & 0.18 & 2.00 & 0.00 & 11.20 &  \multirow{2}{*}{Mann–Whitney ${\displaystyle U}$ test}& \multirow{2}{*}{10461.00} & \multirow{2}{*}{0.219183} \\
    &Enabled & 364 & 1.05 & 0.25 & 2.18 & 0.00 & 20.18 & & & \\ 
    \bottomrule
  \end{tabular}
\end{table*}

An overview of the results is provided in Table \ref{tab:performance_metrics_TrustMHE}. The p-value indicates that TrustMHE has a significant impact on both the \textsc{Crashes} and \textsc{Success} metrics, which are correlated. Furthermore, TrustMHE does not show a significant effect on the \textsc{Progress} metric within the case study. With respect to the \textsc{Min. Dist.} metric, no statistical significance is found. However, this is likely due to the reduced sample size, which was limited to successful test runs. Despite the small statistical sizes, the results indicate that TrustMHE affects \textsc{Min. Dist.}, as well. Overall, it can be concluded that hypothesis \ref{h:0} is rejected, and hypothesis \ref{h:1} is accepted.

\begin{figure*}[]
	\centering	
	\resizebox{1\textwidth}{!}{
        {\scalefont{1.6} \input{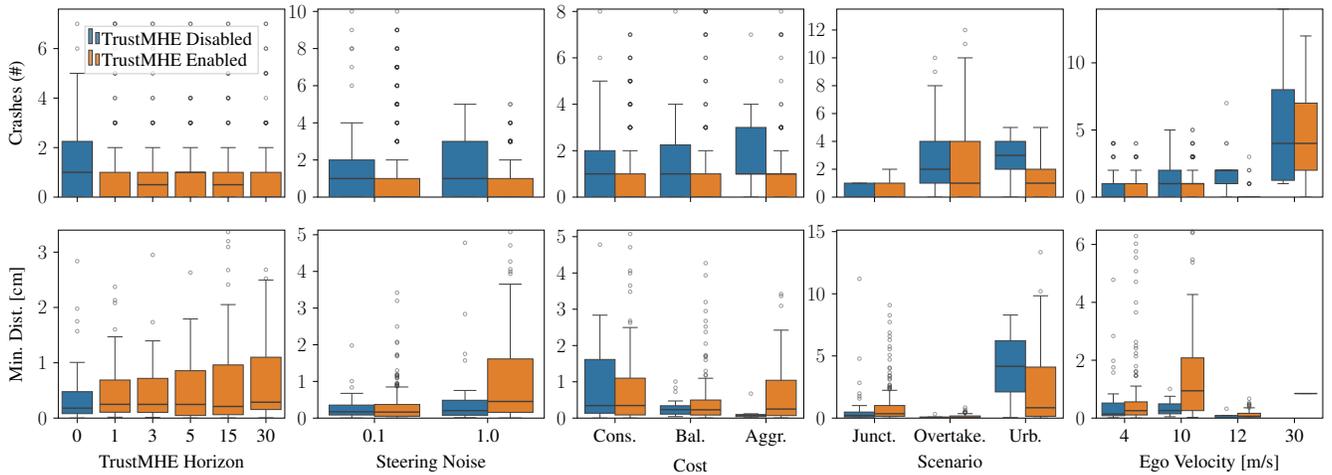}}
    }
	\caption{Evaluation of TrustMHE across different experimental setting variations. Overall TrustMHE "Disabled" is compared against five slightly different TrustMHE "Enabled" settings. In particular, five different horizons $T_{\mathrm{est}}=\{1,3,5,15,30\}$ settings are considered and separately illustrated.}
	\label{fig:Eval2}
\end{figure*}

As shown in Table \ref{tab:performance_metrics_TrustMHE}, five times as many test runs are conducted with TrustMHE set to "Enabled" compared to "Disabled." This is because TrustMHE "Disabled" is compared against five slightly different TrustMHE "Enabled" settings. In particular, five different horizons $T_{\mathrm{est}}=\{1,3,5,15,30\}$ settings are considered. While the tuning parameter's effect is neglected in the overall hypothesis testing (Table \ref{tab:performance_metrics_TrustMHE}), Figure \ref{fig:Eval2} illustrates the effect of the tuning parameter.

Figure \ref{fig:Eval2} presents a comparison of the impact on the significant metric \textsc{Crashes}, and the metric \textsc{Min. Dist.}, considering not only the TrustMHE horizon evaluation but also further insights w.r.t planner variations, e.g., steering sampling noise and cost function modes, as well as scenario-dependent effects like topology and velocities. As shown, the performance of the TrustMHE approach is not substantially affected by the chosen horizon, with a statistical test yielding a p-value of 0.743751. A similar trend is observed with the variation in MPPI sampling steering noise. However, regarding the different planner modes, it becomes evident that crashes can be reduced, particularly when compared to the aggressive planner. Furthermore, it is clear that scenario topology and velocities present varying levels of challenge, but in every case, TrustMHE leads to an improvement.

\section{Discussion \& Conclusion }\label{conclusion}

In this paper, the challenge of integrating AI in a closed-loop within a safety-critical cyber-physical system is considered. In this context, we present a concept that can be integrated natively into modern statistical processing chains horizontally and vertically. The concept is analyzed through a case study on trajectory prediction and planning, proving that the hypothesis of improving safety could be confirmed. It should be emphasized that the hypothesis test is designed to investigate whether the methodology leads to a significant improvement, not on the performance perse, such as achieving zero crashes. Overall, the TrustMHE approach proves effective in enhancing safety. Moreover, the conceptual flexibility allows to meet task-specific requirements while merging engineering concepts from safety, control, and AI. Thus, the concept enables to enhance system self-awareness, and ultimately trust of AI. Simultaneously, this opens up future research questions on a more general methodology that consolidates cross-disciplinary concepts to deploy AI in a reliable and trustworthy manner in real world environments and systems. In parallel, future research should also focus on transitioning to real world systems, with an emphasis on embedded real-time implementations and the generalization of simulation results to open long-tail distributions of real world scenarios.

\bibliographystyle{IEEEtran}
\bibliography{literature}

\end{document}